\documentclass[sigconf]{acmart}

\AtBeginDocument{%
  }

\setcopyright{acmlicensed}
\copyrightyear{2024}
\acmYear{2024}
\acmDOI{XXXXXXX.XXXXXXX}

\acmConference[ICCAD'24]{2024 International Conference on Computer-Aided Design}{October 27--31,
  2024}{New Jersey, USA}

\usepackage{array}
\usepackage{algorithm}
\usepackage{amsmath}
\usepackage{amsfonts}
\usepackage{booktabs}
\usepackage{float}
\usepackage{graphicx}
\usepackage{listings}
\usepackage{mathrsfs}
\usepackage{multirow}
\usepackage{multicol}
\usepackage{mwe}
\usepackage{pifont}
\usepackage{siunitx}
\usepackage{soul}
\usepackage[countmax]{subfloat}
\usepackage{xcolor}

\usepackage{url}

\usepackage{algorithmic}
\usepackage{graphicx}
\usepackage{subfig}
\usepackage{textcomp}
\usepackage{xcolor}

\newcolumntype{P}[1]{>{\centering\arraybackslash}p{#1}}

\copyrightyear{2024}
\acmYear{2024}
\setcopyright{rightsretained}
\acmConference[ICCAD '24]{IEEE/ACM International Conference on Computer-Aided Design}{October 27--31, 2024}{New York, NY, USA}
\acmBooktitle{IEEE/ACM International Conference on Computer-Aided Design (ICCAD '24), October 27--31, 2024, New York, NY, USA}
\acmDOI{10.1145/3676536.3697116}
\acmISBN{979-8-4007-1077-3/24/10}

\begin{document}

\title{EDALearn: A Comprehensive RTL-to-Signoff EDA Benchmark for Democratized and Reproducible ML for EDA Research}

\subtitle{(Invited Paper)}

\author{Jingyu Pan}
\email{jingyu.pan@duke.edu}
\affiliation{%
  \institution{Duke University}
  \city{Durham}
  \state{NC}
  \country{USA}
}

\author{Chen-Chia Chang}
\email{chenchia.chang@duke.edu}
\affiliation{%
  \institution{Duke University}
  \city{Durham}
  \state{NC}
  \country{USA}
}

\author{Zhiyao Xie}
\email{eezhiyao@ust.hk}
\affiliation{%
  \institution{Hong Kong University of Science and Technology}
  \country{Hong Kong}
}

\author{Yiran Chen}
\email{yiran.chen@duke.edu}
\affiliation{%
  \institution{Duke University}
  \city{Durham}
  \state{NC}
  \country{USA}
}

\author{Hai (Helen) Li}
\email{hai.li@duke.edu}
\affiliation{%
  \institution{Duke University}
  \city{Durham}
  \state{NC}
  \country{USA}
}


\begin{abstract}
The application of Machine Learning (ML) in Electronic Design Automation (EDA) for Very Large-Scale Integration (VLSI) design has garnered significant research attention.
Despite the requirement for extensive datasets to build effective ML models, most studies are limited to smaller, internally generated datasets due to the lack of comprehensive public resources.
In response, we introduce EDALearn, the first holistic, open-source benchmark suite specifically for ML tasks in EDA.
This benchmark suite presents an end-to-end flow from synthesis to physical implementation, enriching data collection across various stages.
It fosters reproducibility and promotes research into ML transferability across different technology nodes.
Accommodating a wide range of VLSI design instances and sizes, our benchmark aptly represents the complexity of contemporary VLSI designs.
Additionally, we provide an in-depth data analysis, enabling users to fully comprehend the attributes and distribution of our data, which is essential for creating efficient ML models.
Our contributions aim to encourage further advances in the ML-EDA domain.

\end{abstract}




\maketitle

\begin{figure}[t!]
    \centering
    \includegraphics[width=1.05\linewidth]{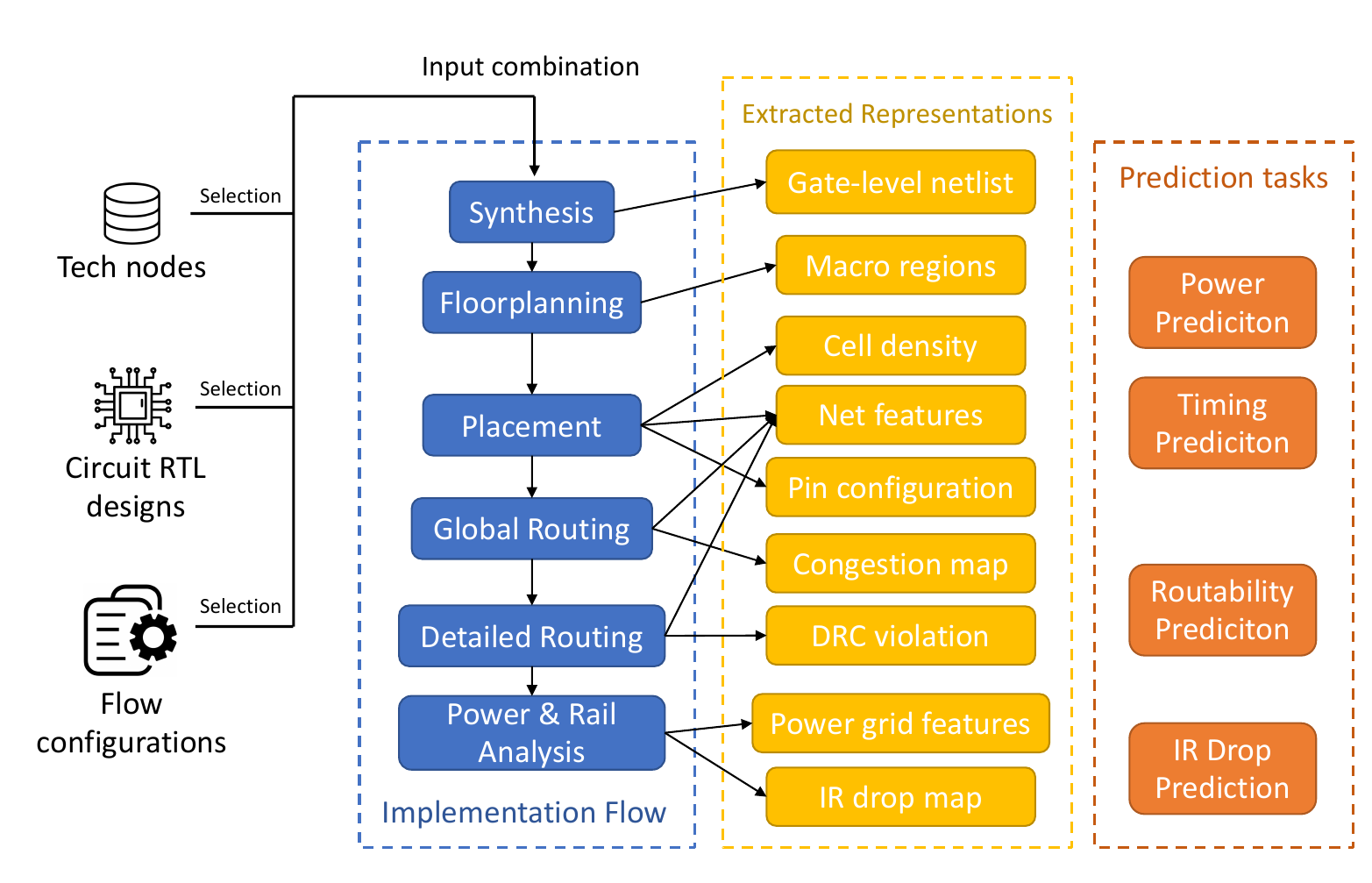}
    \caption{Overview of the EDALearn benchmark.}
    \label{fig:sec1-overview}
\end{figure}

\section{Introduction}

\begin{table*}[t!]
\centering
\caption{Contribution and Limitations of Previous Benchmarks}
\begin{tabular}{|P{2cm}|P{6cm}|P{6cm}|}
\hline
\textbf{Benchmark} & \textbf{Contribution} & \textbf{Limitations} \\ \hline
\multirow{4}{*}{CircuitNet~\cite{chai2022circuitnet}} & \multirow{4}{=}{First open-source dataset for ML applications in EDA.} & \multirow{4}{=}{Insufficient variety of circuits (merely 6 RISC-V designs) and non-reproducible dataset due to use of closed-sourced industrial technology node.} \\
&  &  \\ & & \\ & & \\ \hline
\multirow{2}{*}{TITAN~\cite{murray2013titan}} & \multirow{2}{=}{Focuses on FPGA placement and routing with 23 benchmark circuits.} & \multirow{2}{=}{Limited coverage of possible designs in the VLSI domain.} \\
& & \\ \hline
\multirow{2}{*}{Hutton et al.~\cite{hutton1998characterization}} & \multirow{2}{=}{Discussed parameterized generation of synthetic combinational circuits.} & \multirow{2}{=}{Oversimplified circuit with limited sizes and incomplete methods for sequential circuits.} \\
& &  \\ \hline
\multirow{2}{*}{OpenABC~\cite{chowdhury2021openabc}} & \multirow{2}{=}{Large labeled dataset focusing on the synthesis flow of open-source hardware IPs.} & \multirow{2}{=}{Only focuses on synthesis stage and lacks data collection during backend stages.} \\
& &  \\ \hline
\end{tabular}
\label{tab:contribution}
\end{table*}

The process of VLSI circuit design is typically bifurcated into two stages: front-end and back-end design.
While the former focuses on the circuit's functionality, the latter is responsible for translating the circuit into manufacturable geometries or layouts.
However, as technology continues to advance, back-end design often proves to be a time-intensive process.
This is largely due to the iterative information exchange between the various design stages during the optimization phase.

To mitigate these delays and streamline the process, cross-stage prediction techniques have been introduced.
By replacing long feedback loops between design stages with more localized loops within each stage, these techniques have significantly accelerated the back-end design process.
Machine learning (ML), known for its predictive capabilities, has been widely employed to expedite various early-stage prediction tasks in the design flow, such as routability~\cite{xie2022deep} and IR drop~\cite{huang2021machine}.

In the face of increasing complexity in modern VLSI circuits and the evolving demands of advanced technology nodes, EDA tools are experiencing mounting challenges.
In response to this, there is a growing urgency for more intelligent and efficient methods to enhance the design process.
The objective is to optimize the performance, power, and area (PPA) metrics of VLSI circuits.
In this respect, machine learning stands out as a promising solution with its data-driven approach and prediction capabilities, providing invaluable assistance in overcoming challenges in the EDA domain.

This paper presents EDALearn\footnote{The dataset and a reference flow of our EDALearn benchmark suite is open-sourced at \url{https://github.com/panjingyu/EDALearn}.}, a comprehensive benchmark suite for evaluating and comparing ML-based approaches in the context of EDA.
Figure~\ref{fig:sec1-overview} presents the overview of the EDALearn benchmark flow.
The benchmark covers a wide range of critical EDA tasks, such as placement, routing, power analysis, timing analysis, and IR drop prediction.
The suite includes a diverse set of real-world VLSI design instances, along with a standardized evaluation framework that facilitates the fair assessment of different ML models and techniques.

By providing a systematic and unified platform for performance evaluation, the proposed EDALearn benchmark aims to foster collaboration and knowledge sharing within the EDA community, encourage the development of novel ML-based algorithms and methodologies, and ultimately contribute to the advancement of EDA tools and processes.
Our contribution is listed as follows:
\begin{itemize}
    \item We introduce a holistic benchmark suite called EDALearn which provides an end-to-end flow from synthesis to physical implementation, and present comprehensive and detailed data collection across multiple stages of our flow, thus addressing the limitations of existing benchmarks that only focus on certain stages.
    \item We provide an open-source reference flow for our EDALearn benchmark, fostering reproducibility and enabling research on ML transferability across different technology nodes, an area overlooked by other benchmarks~\cite{chai2022circuitnet}.
    \item Unlike benchmarks limited in their coverage of possible designs and circuit sizes~\cite{murray2013titan,hutton1998characterization,chai2022circuitnet}, our EDALearn benchmark accommodates a wide variety of VLSI design instances and sizes, making it more representative of modern VLSI design complexity.
    \item We provide an in-depth data analysis, enabling users to better understand the attributes and distribution of our data, which is critical for developing effective ML models.
\end{itemize}

The remainder of this paper is organized as follows:
Section \ref{sec:related-works} summarizes previous ML for EDA benchmarks, highlighting their limitations for comprehensive ML evaluation.
Section \ref{sec:benchmark} describes our EDALearn benchmark suite in detail, including the used design instances, technology nodes, flow variations, extracted feature representations, and supported EDA tasks.
Section \ref{sec:analysis} presents a data analysis of the EDALearn benchmark.
Finally, Section \ref{sec:conclusion} concludes the paper and outlines future ML for EDA research directions enabled by our EDALearn benchmark.

\section{Related Works}
\label{sec:related-works}

In this seciton, we review previous ML for EDA benchmarks and discuss their limitations for comprehensive ML evaluation.

CircuitNet~\cite{chai2022circuitnet} is the first open-source dataset for machine learning applications in electronic design automation (EDA).
It provides holistic support for cross-stage prediction tasks in back-end design with diverse samples.
However, it only adopted 6 RISC-V designs in total, which is very limited for practical ML for EDA tasks.
CircuitNet2~\cite{jiang2024circuitnet2}, an updated version of CircuitNet, included some GPU/AI chip designs to increase the span of the designs and covered a 14nm FinFET tech node besides the 28nm CMOS tech node in the earlier version.
Both are based on closed-sourced industrial technology nodes, making the generated dataset not reproducible.

TITAN~\cite{murray2013titan} is a benchmark that focuses on FPGA placement and routing.
It uses Altera’s Quartus II FPGA CAD software to perform HDL synthesis and creates the Titan23 benchmark set, which consists of 23 benchmark circuits.
However, the coverage of these circuits is limited to a narrow range of possible designs or use cases in the general VLSI domain.
\cite{elgammal2021rlplace} used the TITAN benchmark to demonstrate a reinforcement learning (RL)-based optimization for FPGA placement.

\cite{hutton1998characterization} discussed parameterized generation of synthetic combinational circuits, providing a convenient way to generate large numbers of circuits with a wide range of characteristics.
However, the sizes of the generated circuits are limited to a few thousand gates, which is insufficient for modern VLSI design.
Besides, the paper does not extend its methodology to the generation of sequential circuits, limiting its applicability in comprehensively simulating and testing the full spectrum of modern VLSI designs.

\begin{figure*}
    \centering
    \subfloat[Cell density.]{\includegraphics[width=0.18\textwidth]{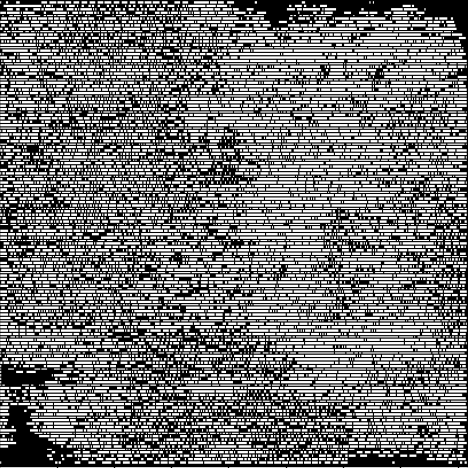}\label{fig:feature-exmaple-1a}}
    \hfill
    \subfloat[Density of CTS cells.]{\includegraphics[width=0.18\textwidth]{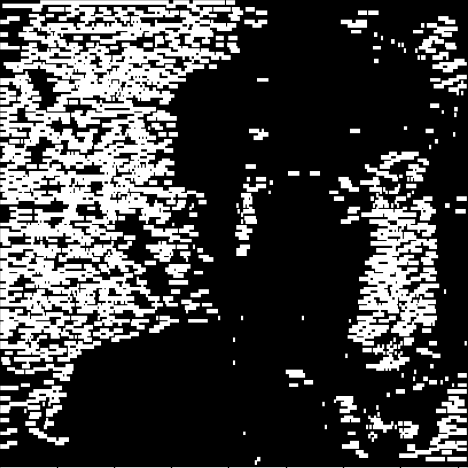}\label{fig:feature-exmaple-1b}}
    \hfill
    \subfloat[Pin density.]{\includegraphics[width=0.18\textwidth]{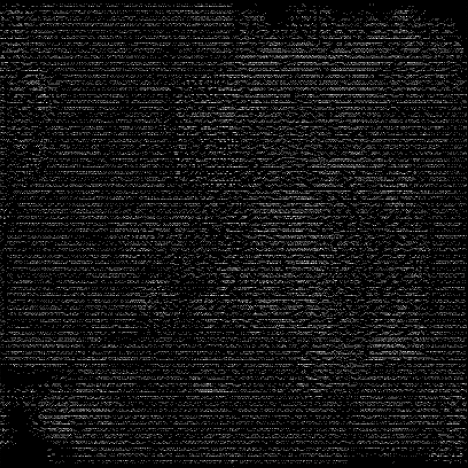}\label{fig:feature-exmaple-1c}}
    \hfill
    \subfloat[Pin accessibility~\cite{yu2019pin}.]{\includegraphics[width=0.18\textwidth]{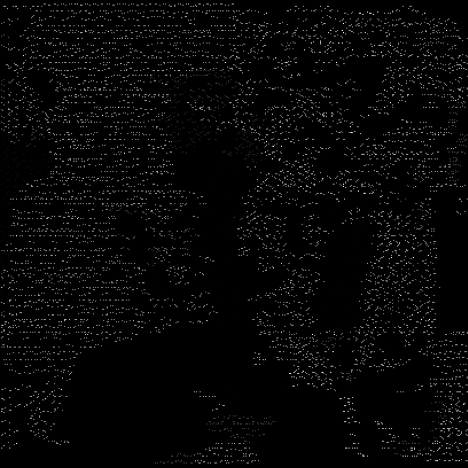}\label{fig:feature-exmaple-1d}}
    \hfill
    \subfloat[Pin density of CTS cells.]{\includegraphics[width=0.18\textwidth]{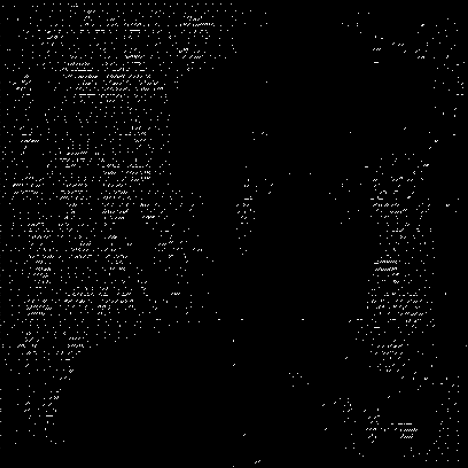}\label{fig:feature-exmaple-1e}}
    \\
    \subfloat[Bounding box of cells.]{\includegraphics[width=0.18\textwidth]{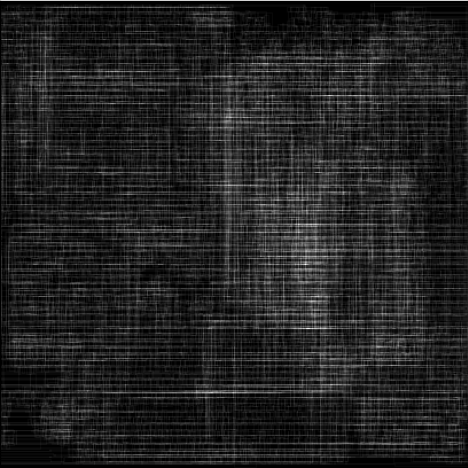}\label{fig:feature-exmaple-2a}}
    \hfill
    \subfloat[RUDY~\cite{spindler2007fast}.]{\includegraphics[width=0.18\textwidth]{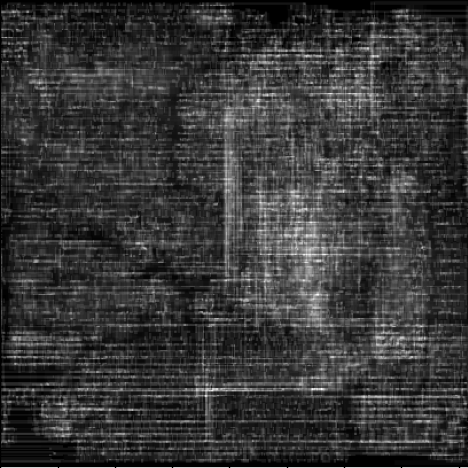}\label{fig:feature-exmaple-2b}}
    \hfill
    \subfloat[Horizontal net density.]{\includegraphics[width=0.18\textwidth]{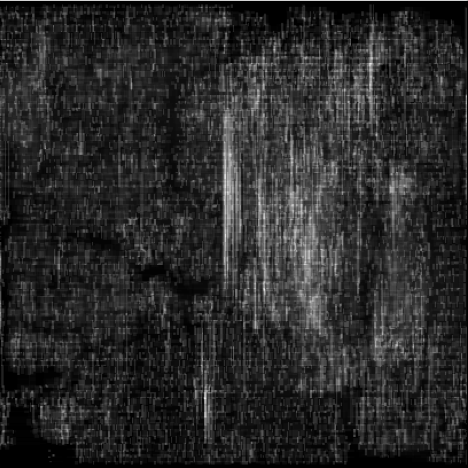}\label{fig:feature-exmaple-2c}}
    \hfill
    \subfloat[Vertical net density~\cite{chen2020pros}.]{\includegraphics[width=0.18\textwidth]{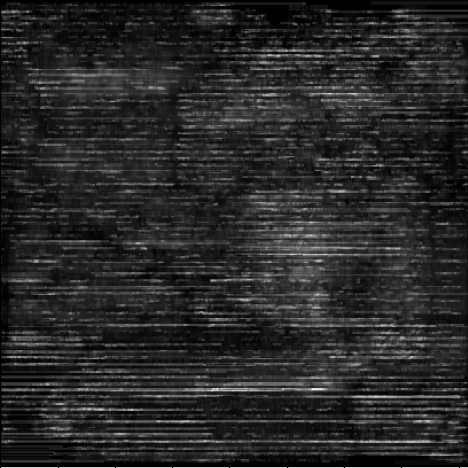}\label{fig:feature-exmaple-2d}}
    \hfill
    \subfloat[Fly lines of all nets.]{\includegraphics[width=0.18\textwidth]{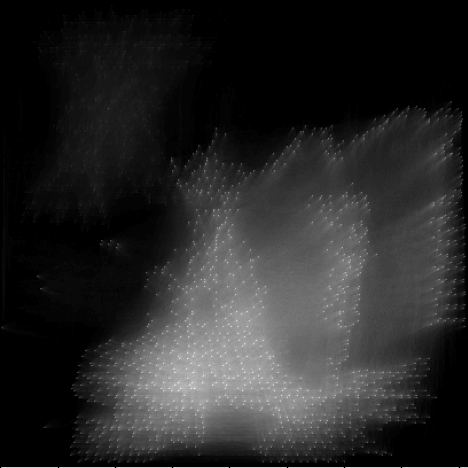}\label{fig:feature-exmaple-2e}}
    \\
    \subfloat[Fly lines of nets (V1).]{\includegraphics[width=0.18\textwidth]{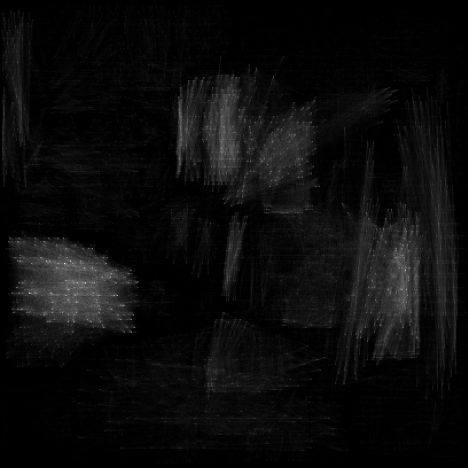}\label{fig:feature-exmaple-3a}}
    \hfill
    \subfloat[Fly lines of nets (V2).]{\includegraphics[width=0.18\textwidth]{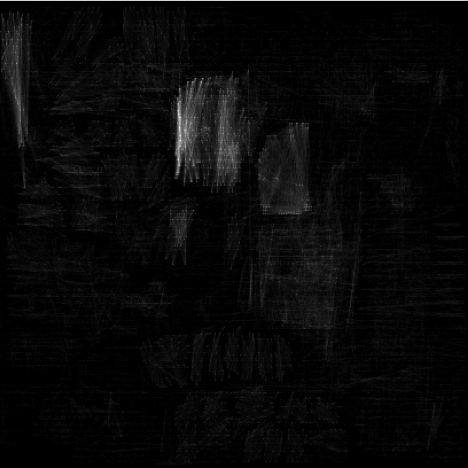}\label{fig:feature-exmaple-3b}}
    \hfill
    \subfloat[Fly lines of nets (V3).]{\includegraphics[width=0.18\textwidth]{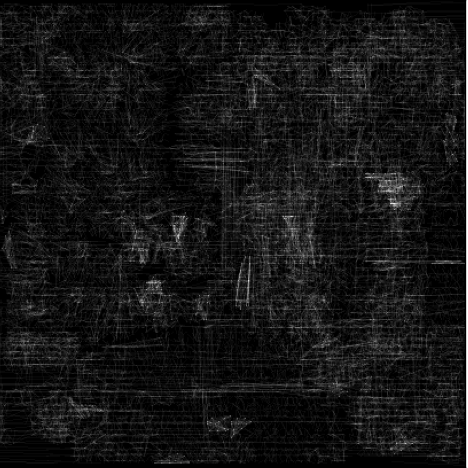}\label{fig:feature-exmaple-3c}}
    \hfill
    \subfloat[Static IR drop.]{\includegraphics[width=0.18\textwidth]{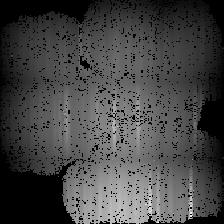}\label{fig:feature-exmaple-3d}}
    \hfill
    \subfloat[DRC violation hotspots.]{\includegraphics[width=0.18\textwidth]{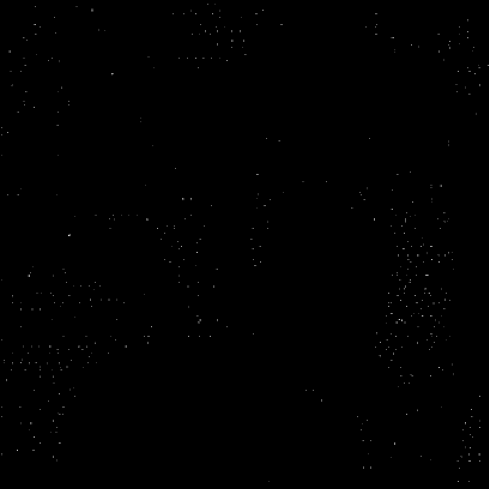}\label{fig:feature-exmaple-3e}}
    \caption{Examples of image-like feature and label representations. "CTS cells" refers to cells directly connected to clock tree nets. Sub-figures (k)-(n) depict varying net selections, denoted as V1-V3, each based on a different bounding box area threshold: 4k $\mu$m for V1, 1k $\mu$m for V2 and 0.1k$\mu$m for V3. Two examples of post-routing stage labels are IR drop maps and DRC violation hotspots.}
    \label{fig:feature-examples}
\end{figure*}

OpenABC~\cite{chowdhury2021openabc} is also a large labeled dataset whose focus is the synthesis flow of open-source hardware IPs.
The hardware IP designs are collected from various sources, including MIT\_CEP, IWLS, OpenROAD, OpenPiton, etc..
For each hardware IP, it generates synthesis results with around 1500 random variations, or so-called synthesis recipes.
\cite{zhou2023area} explored RL-based area optimization of logic synthesis on FPGAs based on the data collected from OpenABC.

While contributions such as CircuitNet, TITAN, synthetic combinational circuits, and OpenABC have set the stage for the integration of machine learning applications in EDA, they also exhibit certain shortcomings.
These limitations range from restricted design diversity, reliance on closed-sourced technology nodes, limited size of generated circuits, to a narrowed focus on specific stages of the design process.
Most importantly, current benchmarks lack comprehensive ML evaluation capabilities, which is instrumental in addressing practical EDA challenges.
Despite these challenges, each benchmark has added value to the broader pursuit of refining ML for EDA, paving the way for future research.
Moving forward, the integration of a more comprehensive, diverse, and reproducible benchmark suite would be instrumental in mitigating these limitations, advancing the scope and precision of machine learning applications in EDA.
This paper introduces such a benchmark, designed to harness the strengths of the earlier works while addressing their limitations, offering an advanced framework for evaluating ML models in the EDA domain.

\section{The EDALearn Benchmark}
\label{sec:benchmark}

This section presents the details of our comprehensive benchmark suite, designed to evaluate and compare ML-based approaches in the context of EDA tasks.
The benchmark suite is based on the FreePDK 45nm~\cite{stine2007freepdk}, the Synopsys Armenia Educational Department (SAED) 32nm~\cite{goldman201332}, and ASAP 7nm~\cite{clark2016asap7} technology nodes, and it supports routability prediction, IR drop prediction, and cross-stage data analysis.
The suite includes design instances from the IWLS benchmark and BOOM CPU designs from Chipyard, facilitating a diverse and practical evaluation of ML models and techniques.

\begin{table}[tb]
  \centering
  \caption{Synthesis variations.}
  \begin{tabular}{c@{\quad}c@{\quad}c}
    \toprule
    \multicolumn{2}{c}{\textbf{Parameters}} & \textbf{Values} \\
    \midrule
    \multicolumn{2}{c}{Frequency (MHz)} & 100/500/1000 \\
    \midrule
    \multirow{3.5}{*}{DRC} & Max Fanout & 10/40 \\
    & Max Transition & 0.5 \\
    & Max Capacitance & 5 \\
    \midrule
    \multicolumn{2}{c}{Command} & compile/compile\_ultra \\
    \midrule
    \multirow{2.5}{*}{Options} & Dynamic Power Optimization & on/off \\
    & Leakage Power Optimization & on/off \\
    \bottomrule
  \end{tabular}
  \label{tab:syn-variations}
\end{table}

\begin{table}[tb]
  \centering
  \caption{Physical design variations (global options).}
  \begin{tabular}{c@{\quad}c}
    \toprule
    \textbf{Parameters} & \textbf{Values} \\
    \midrule
    Design Flow Effort & express/standard \\
    Design Power Effort & none/high \\
    Target Utilization (\%) & 50/60/70/80/90 \\
    \bottomrule
  \end{tabular}
  \label{tab:pd-variations-global}
\end{table}

\begin{table}[tb]
  \centering
  \caption{Physical design variations (placement options).}
  \begin{tabular}{c@{\quad}c}
    \toprule
    \textbf{Parameters} & \textbf{Values} \\
    \midrule
    Global Timing Effort & medium/high \\
    Global Congestion Effort & medium/high \\
    Detail Wire Length Opt Effort & medium/high \\
    Global Max Density & 0.75/0.9 \\
    Activity Power Driven & on/off \\
    \bottomrule
  \end{tabular}
  \label{tab:pd-variations-placement}
\end{table}

\begin{table}[tb]
  \centering
  \caption{Physical design variations (CTS options).}
  \begin{tabular}{c@{\quad}c}
    \toprule
    \textbf{Parameters} & \textbf{Values} \\
    \midrule
    Pre-CTS Opt Max Density & 0.8 \\
    Pre-CTS Opt Power Effort & none/low/high \\
    Pre-CTS Opt Reclaim Area & on/off \\
    Pre-CTS Fix Fanout Load & on/off \\
    \midrule
    Cell Density & 0.5 \\
    Clock Gate Buffering Location & below \\
    Clone Clock Gates & on \\
    \midrule
    Post-CTS Opt Power Effort & low/high \\
    Post-CTS Opt Reclaim Area & on/off \\
    Post-CTS Fix Fanout Load & on/off \\
    \bottomrule
  \end{tabular}
  \label{tab:pd-variations-cts}
\end{table}

\subsection{Technology Nodes}


Our benchmark suite covers 3 distinct technology nodes: FreePDK 45nm, SAED 32nm and ASAP 7nm.
All of them are open-sourced tech nodes, making our data fully-reproducible.
Each node brings its unique attributes and characteristics to the fore, providing a diverse and challenging landscape for the assessment of machine learning techniques within the sphere of EDA tasks.

The FreePDK 45nm and SAED 32 nodes represents an earlier phase in semiconductor manufacturing technology, characterized by larger feature sizes and lower transistor density. Consequently, circuits designed at this node tend to be more straightforward with respect to layout and thermal management, yet pose challenges in terms of power efficiency and high-speed performance.

On the other hand, the ASAP 7nm node exemplifies the cutting-edge in semiconductor technology, characterized by extremely small feature sizes and high transistor density. Designs at this node face significant challenges related to power, performance, and area trade-offs, as well as manufacturing yield and reliability. Furthermore, the high complexity and cost associated with this advanced node necessitate efficient and accurate EDA tools to ensure optimal design outcomes.

Through the juxtaposition of these distinct technology nodes, our benchmark suite offers a broad perspective, enabling a comprehensive and insightful evaluation of ML techniques across diverse technological contexts in EDA tasks.

\subsection{Flow Variations}

Table~\ref{tab:syn-variations} lists the synthesis variations used in the benchmark suite.
Table~\ref{tab:pd-variations-global}, Table~\ref{tab:pd-variations-placement}, and Table~\ref{tab:pd-variations-cts} shows the global option variations, placement-stage variations and CTS-stage variations used for the physical design, respectively.
The inclusion of these variations affords a broad spectrum of potential physical implementations.
This diversification in data distribution significantly enhances the scope for development in machine learning for EDA by accommodating an array of possible real-world scenarios.

\subsection{Design Instances}


Our EDALearn benchmark suite comprises a varied selection of practical VLSI design instances, promoting a thorough and dependable assessment of machine learning models and methodologies.
Table~\ref{tab:design-statistics} shows the statistics of the design instances, each sourced from an openly accessible benchmark, offers a diverse range of circuit statistics.
We proceed to detail the benchmarks that serve as the origin of these designs.

\begin{table}[tb]
  \centering
  \caption{Circuit statistics of representative designs.}
  \begin{tabular}{ccccc}
    \toprule
    \multirow{2.5}{*}{\textbf{Designs}} & \multicolumn{3}{c}{\textbf{Circuit Statistics}} & \multirow{2.5}{*}{\textbf{Source}} \\
    \cmidrule(lr){2-4}
    & \#Cells & \#Nets & \#Macros & \\
    \midrule
    Small BOOM & 686k & 898k & 39 & Chipyard \\
    Medium BOOM & 998k & 1247k & 39 & Chipyard \\
    simple\_spi & 500 & 800 & 0 & OpenCores \\
    pci & 25565 &  33838 & 0 & OpenCores \\
    ac97\_ctrl & 11317 & 14032 & 0 & OpenCores \\
    b14 & 19676 & 21983 & 0 & ITC'99 \\
    b14\_1 & 19622 & 21912 & 0 & ITC'99 \\
    b15 & 10019 & 11460 & 0 & ITC'99 \\
    b15\_1 & 10017 & 11455 & 0 & ITC'99 \\
    b17 & 30563 & 35197 & 0 & ITC'99 \\
    b17\_1 & 30697 & 35312 & 0 & ITC'99 \\
    b18 & 87803 & 103k & 0 & ITC'99 \\
    b18\_1 & 86107 & 102k & 0 & ITC'99 \\
    b19 & 173k & 205k & 0 & ITC'99 \\
    b20 & 38511 & 43607 & 0 & ITC'99 \\
    b20\_1 & 38584 & 43667 & 0 & ITC'99 \\
    b21 & 38363 & 43427 & 0 & ITC'99 \\
    b21\_1 & 39886 & 44978 & 0 & ITC'99 \\
    b22 & 56931 & 64478 & 0 & ITC'99 \\
    b22\_1 & 57469 & 65046 & 0 & ITC'99 \\
    des & 2895 & 3388 & 0 & OpenCores \\
    DMA & 29268 & 39715 & 0 & Faraday \\
    DSP & 47663 & 64068 & 0 & Faraday \\
    ethernet & 69546 & 82199 & 0 & OpenCores \\
    fpu & 34039 & 37226 & 0 & OpenCores \\
    i2c & 663 & 778 & 0 & OpenCores \\
    leon2 & 699k & 755k & 0 & Gaisler \\
    mem\_ctrl & 10015 & 12212 & 0 & OpenCores \\
    pci & 25565 & 33838 & 0 & OpenCores \\
    RISC & 81160 & 110k & 0 & Faraday \\
    s1196 & 578 & 597 & 0 & ISCAS \\
    s1238 & 623 & 643 & 0 & ISCAS \\
    s13207 & 1703 & 1844 & 0 & ISCAS \\
    s1423 & 989 & 1068 & 0 & ISCAS \\
    s832 & 424 & 449 & 0 & ISCAS \\
    s838\_1 & 363 & 431 & 0 & ISCAS \\
    s9234\_1 & 1338 & 1479 & 0 & ISCAS \\
    sasc & 786 & 909 & 0 & OpenCores \\
    spi & 2890 & 3285 & 0 & OpenCores \\
    ss\_pcm & 433 & 511 & 0 & OpenCores \\
    systemcaes & 10624 & 12617 & 0 & OpenCores \\
    systemcdes & 3693 & 4376 & 0 & OpenCores \\
    tv80 & 7724 & 8378 & 0 & OpenCores \\
    usb\_funct & 18845 & 23913 & 0 & OpenCores \\
    usb\_phy & 681 & 794 & 0 & OpenCores \\
    vga\_lcd & 89751 & 105k & 0 & OpenCores \\
    wb\_conmax & 64874 & 99445 & 0 & OpenCores \\
    wb\_dma & 81615 & 94562 & 0 & OpenCores \\
    \bottomrule
  \end{tabular}
  \label{tab:design-statistics}
\end{table}

\begin{itemize}

  \item \textbf{OpenCore~\cite{opencores}:} A set of open-source hardware designs from the OpenCore project. These designs cover a broad spectrum of complexity and are ideal for evaluating the effectiveness of ML techniques in various open-source hardware scenarios.

  \item \textbf{ISCAS~\cite{brglez1985neutral,brglez1989combinational}:} A collection of benchmark circuits from the International Symposium on Circuits and Systems (ISCAS). These designs, which include both combinational and sequential circuits, provide a diverse set of challenges for assessing ML techniques in the context of circuit analysis and optimization.

  \item \textbf{Gaisler~\cite{gaisler}:} A series of designs from Gaisler Research, known for their work in digital hardware design for commercial and aerospace applications. These designs, including the LEON2 and LEON3 processors, offer a unique opportunity to evaluate ML techniques in the context of both commercial and space-grade hardware design.

  \item \textbf{Faraday~\cite{faraday}:} A set of designs from Faraday Technology Corporation, a leading ASIC design service and IP provider. These designs represent a range of modern ASIC implementations, providing a robust platform for assessing ML techniques in the context of state-of-the-art ASIC design.

  \item \textbf{ITC'99\cite{corno2000rt}:} A collection of designs from the International Test Conference (ITC) 1999 benchmark suite. These designs, which cover a wide range of complexity and design styles, are ideal for evaluating the effectiveness of ML techniques in various test and diagnosis scenarios.

  \item \textbf{BOOM CPU designs from Chipyard~\cite{chipyard}:} A set of modern, high-performance RISC-V CPU designs from the Chipyard framework. These designs represent state-of-the-art VLSI implementations and provide an excellent testbed for evaluating ML approaches in the context of advanced designs.

\end{itemize}

\subsection{Circuit Features}

A critical aspect of applying ML techniques in the EDA domain lies in the ability to represent circuit designs in a format amenable to learning. Our benchmark suite offers a comprehensive set of circuit features, encapsulating the necessary information about the circuits into two primary forms: image-like representations and vector-based features.

\begin{itemize}

\item \textbf{Image-like representations:} These features capture the spatial distribution of various circuit components on the silicon area. The layout of a circuit can significantly influence the performance, power, and area (PPA) metrics of a VLSI design. For instance, closely packed gates can lead to higher power consumption and may affect the routability. Figure~\ref{fig:feature-examples} showcases some examples of the image-like feature representations generated based on our EDALearn benchmark. Their representations capture the spatial layout and density of cells, nets, and macros in the design. These features can be easily stacked as tensors to be ingested by convolutional neural network (CNN) models, making them particularly suitable for tasks such as routability prediction, congestion estimation, and IR drop prediction.

\item \textbf{Vector-based features:} These features capture circuit statistics and other design attributes as multi-dimensional vectors. They include parameters like the total number of cells, number of nets, design hierarchy depth, number and type of macros, target clock frequency, among others. These vector-based features serve as abstract representations of the designs and can be utilized by various types of ML models including decision trees, support vector machines, and fully connected neural networks. They are particularly useful for tasks that require a high-level understanding of the design, such as power and timing prediction.

\end{itemize}

By combining both spatial and statistical features, our EDALearn benchmark offers a holistic view of the circuits, enabling ML models to exploit both global and local information in their predictions. Moreover, the feature extraction process is flexible and extensible, allowing researchers to add more features that they believe may be relevant for their specific tasks. This hybrid approach paves the way for exploring complex relationships between circuit attributes and design outcomes, fostering more insightful and accurate ML models for EDA.






\subsection{EDA Tasks}
Our EDALearn benchmark suite covers the following critical EDA tasks:

\begin{itemize}

  \item \textbf{Power prediction:} Estimating the power consumption of a circuit design is a critical task in EDA. Excessive power consumption can lead to overheating, decreased reliability, and shorter battery life in mobile devices. ML models can be used to predict power consumption early in the design process, allowing designers to make necessary modifications to meet power constraints. This task specifically involves predicting both dynamic and leakage power for a given design, considering the impact of factors such as operating voltage, clock frequency, and gate capacitance.

  \item \textbf{Slack time prediction:} Timing analysis is a vital part of the EDA process to ensure that a circuit meets its performance targets. Slack time is the total time that you can delay a task without delaying the project. More slack time indicates a higher chance of meeting the timing requirements even with variations in gate delays. ML models can predict the slack time for each timing path, enabling designers to identify potential timing violations and make necessary adjustments early in the design process. Specifically, this task involves predicting the slack time at different stages of the design flow, such as post-synthesis, post-placement, and post-routing.

  \item \textbf{Routability prediction:} Assessing the feasibility of routing a design within given constraints, such as routing resources and timing requirements. ML models can be employed to predict routability early in the design process, reducing the need for time-consuming iterative refinements. Specifically, we extract labels for routing congestion and post-routing DRC violations, which are the two main metrics for routability.

  \item \textbf{IR drop prediction:} Estimating the voltage drop across the power distribution network, which can impact the performance and reliability of a design. ML-based approaches can provide accurate IR drop predictions in the early stages of design, allowing for more informed power optimization decisions.
\end{itemize}

\begin{figure}[b]
    \centering
    \includegraphics[width=\linewidth]{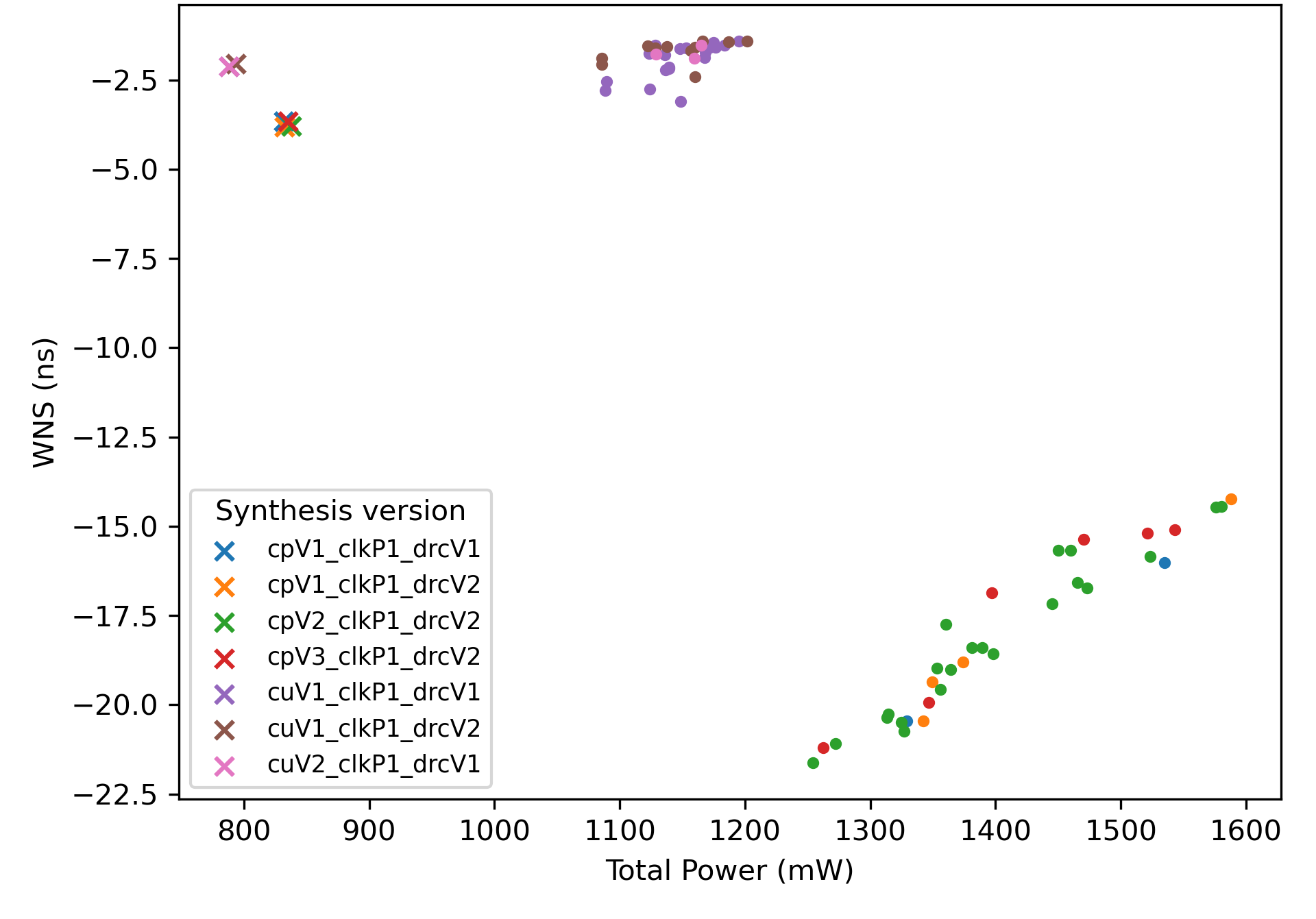}
    \caption{Distribution of WNS and total power from different synthesis/implementation results of Medium BOOM CPU design. Here we compare results based on both the \texttt{compile} and \texttt{compile\_ultra} synthesis command.}
    \label{fig:mediumboom-dc-innovus-scatter}
\end{figure}

In addition to the aforementioned EDA tasks, our EDALearn benchmark suite enables the benefits of cross-stage data analysis.
This analysis involves comparing estimations generated by EDA tools at different stages of the design flow, such as power, timing, or routability estimations at synthesis, placement, and post-routing stages.
By offering insights into the discrepancies and potential areas for improvement in the design flow, cross-stage data analysis can aid in the development of ML models tailored for EDA problems, leading to better optimization and more accurate predictions.

\section{Benchmark Analysis}
\label{sec:analysis}

This section conducts an analysis of the data distribution observed in the EDALearn benchmark.
Specifically, it compares the power and timing data at different stages of the implementation flow.
Furthermore, it extends these comparisons across different technology nodes.
The aim of these analyses is to shed light on potential research directions enabled by the EDALearn benchmark.

\begin{figure}[t]
    \centering
    \includegraphics[width=\linewidth]{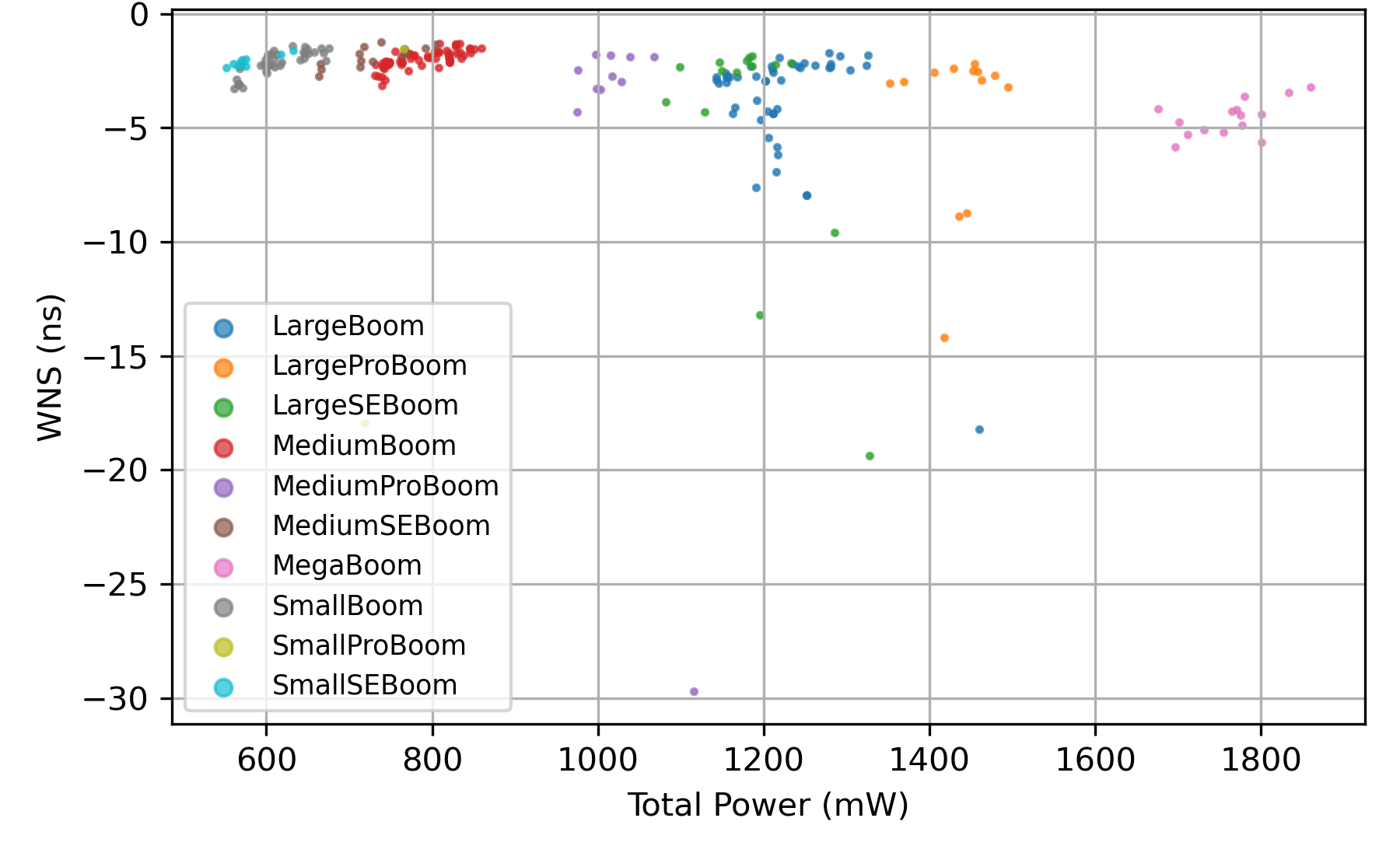}
    \caption{WNS and total power distribution of chipyard Boom designs implemented based on FreePDK45.}
    \label{fig:wns-power-chipyard-freepdk45}
\end{figure}

\subsection{Trade-off between timing and power}

Figure~\ref{fig:mediumboom-dc-innovus-scatter} shows the distribution of worst negative slack and total power from different synthesis/implementation results of Medium BOOM CPU design from chipyard.
Here we compare the difference of distribution observed from results of the \texttt{compile} and \texttt{compile\_ultra} synthesis command.
It clearly shows that the results from both synthesis commands form two distinct clusters, indicating a noticeable difference in the outcomes based on the command used.
The \texttt{compile} command results in one distribution, while the \texttt{compile\_ultra} command results in another.
Most notably, the \texttt{compile\_ultra} command appears to produce superior results in terms of both worst negative slack time and total power consumption.
It is found that the \texttt{compile\_ultra} command results in less worst negative slack time, indicating a better timing performance.
Similarly, the total power consumption is less with the \texttt{compile\_ultra} command, suggesting improved energy efficiency.


\subsection{Chipyard Boom designs}

The discussion here centers on the distribution of Chipyard Boom CPU designs, which bear distinct attributes due to their larger size and incorporation of memory macros.
As depicted in Figure~\ref{fig:wns-power-chipyard-freepdk45}, the worst negative slack time and total power of the Boom CPU designs, based on the FreePDK 45nm technology node, are distributed in unique clusters corresponding to each design size.
An observable trend is that the total power consumed is commensurate with the size of the design, presenting a logical correlation between these two parameters.
When scrutinizing the small and medium Boom CPUs, it becomes apparent that their worst negative slack times are notably similar.
However, this isn't the case for the larger designs.
For the large Boom CPUs and Mega Boom CPUs, the worst negative slack time is conspicuously higher, suggesting an increased complexity in managing timing for these larger designs.
This analysis provides valuable insights into the intricate interplay between design size, power consumption, and timing performance.

\begin{figure}[b]
    \vspace{-2em}
    \centering
    \includegraphics[width=\linewidth]{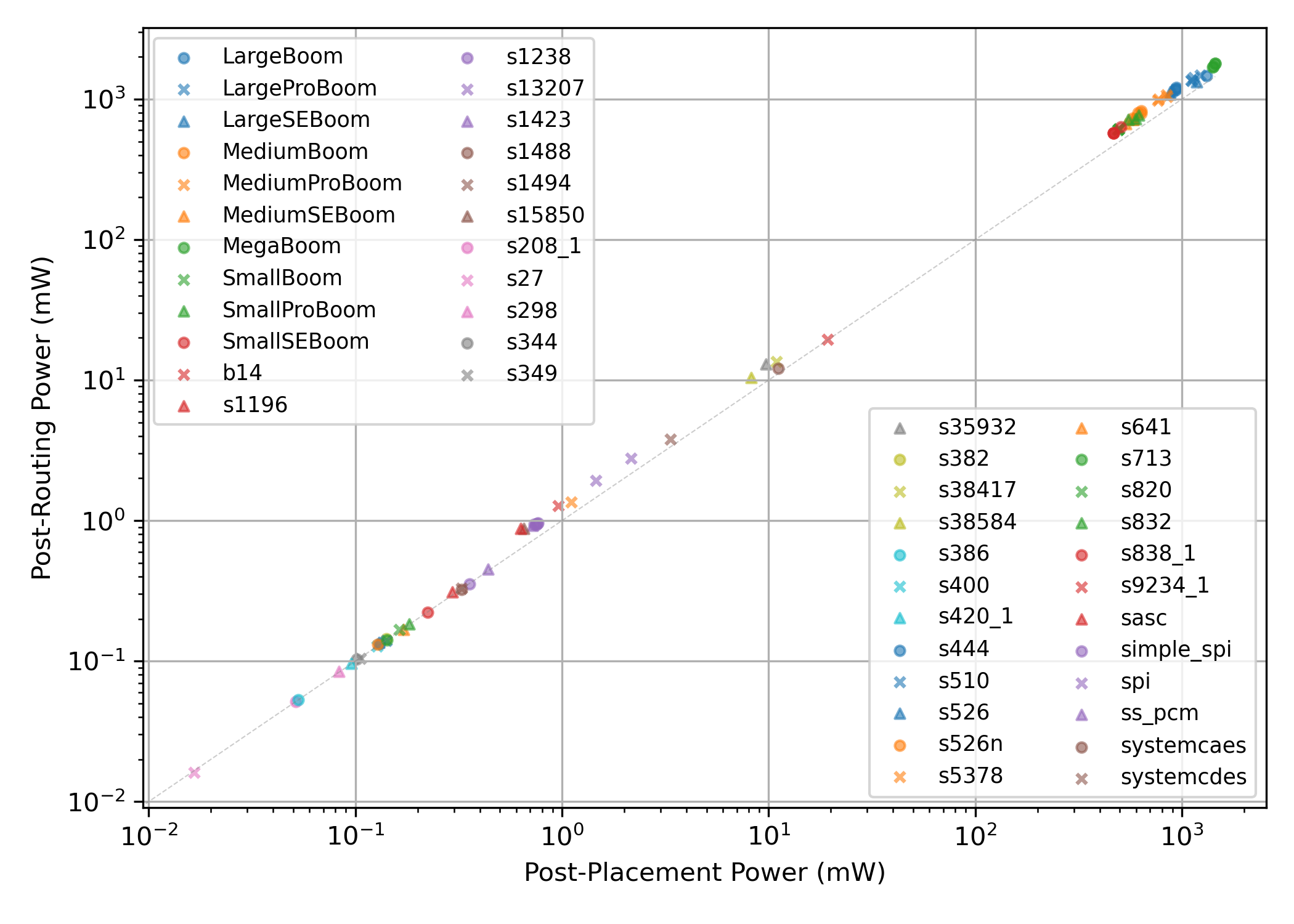}
    \caption{Total power distribution comparison between post-placement stage and post-routing stage of implementations based on the FreePDK 45nm technology node.}
    \label{fig:power-placement-routing-freepdk45}
\end{figure}

\begin{figure}[t]
    \centering
    \includegraphics[width=\linewidth]{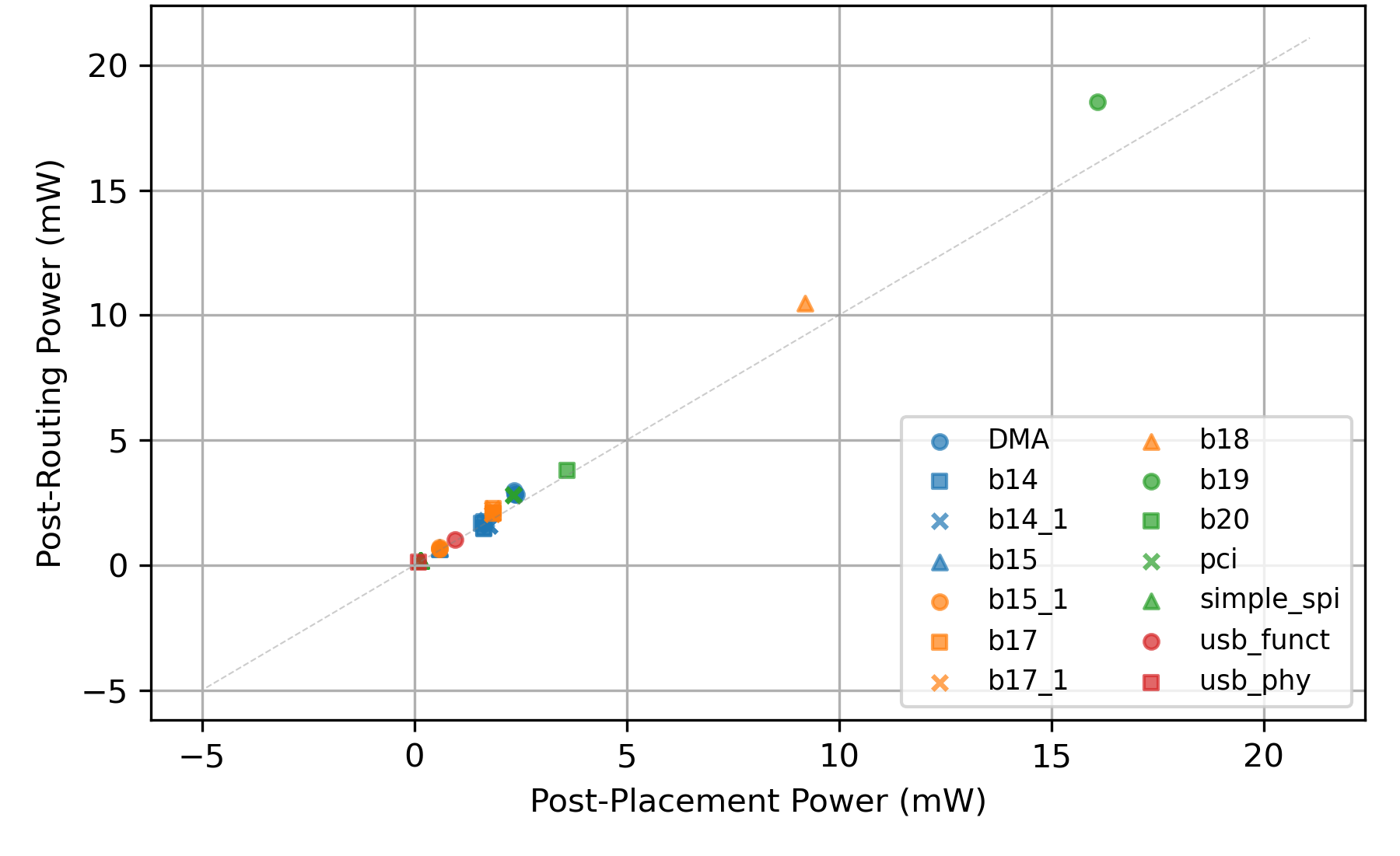}
    \caption{Total power distribution comparison between post-placement stage and post-routing stage of implementations based on the ASAP 7nm technology node.}
    \label{fig:power-placement-routing-asap7}
\end{figure}

\subsection{Comparison between different stages or technology node}

Figure~\ref{fig:power-placement-routing-freepdk45} and Figure~\ref{fig:power-placement-routing-asap7} respectively depict the distribution of total power consumption estimated at the post-routing stage for the FreePDK 45nm and ASAP 7nm technology nodes.
A trend observed in both nodes is that the post-routing total power typically surpasses the post-placement total power, suggesting an over-optimistic estimation of total power at the post-placement stage by the golden tool.
This increase can be attributed to the additional wire length incurred during the routing and optimization stage, which is commonly underestimated.
Notably, the extent of this over-estimation varies between different designs, as evidenced by Figure~\ref{fig:power-placement-routing-freepdk45} and Figure~\ref{fig:power-placement-routing-asap7}.
This observation illustrates the value of our EDALearn benchmark in enabling novel research directions, such as exploring more accurate power estimation models at the post-placement stage, identifying the sources of over-optimism in current tools, or developing ML models capable of more accurately predicting the impact of routing on power consumption.

In addition to offering insights into power estimations, Figure~\ref{fig:power-placement-routing-freepdk45} and Figure~\ref{fig:power-placement-routing-asap7} also facilitate a comparative view of the power distribution differences across two distinct technology nodes.
This comparison not only underscores the diversity in power consumption patterns but also highlights the potential for disparities between different technology nodes.
This helps guide future research endeavors into the transferability of machine learning models, investigating whether models developed and trained on data from one technology node can perform accurately when applied to data from another node.
This would further our understanding of how machine learning models can adapt to new scenarios, potentially leading to more robust and versatile ML solutions in electronic design automation.
\section{Conclusion}
\label{sec:conclusion}


In this work, we have introduced EDALearn, a comprehensive benchmark suite designed to facilitate the evaluation and comparison of ML-based methodologies within the realm of Electronic Design Automation. EDALearn addresses several limitations in existing benchmarks by providing a holistic, end-to-end flow that covers various critical EDA tasks including placement, routing, power analysis, timing analysis, and IR drop prediction.
Our open-source reference flow fosters reproducibility and promotes research on ML transferability across different technology nodes, filling the gap left by previous benchmarks. Furthermore, EDALearn accommodates a broad variety of VLSI design instances and sizes, overcoming the narrow focus of prior benchmarks and making it a more representative solution for the modern, complex VLSI design scenario.
The in-depth data analysis included in the suite enables users to gain a better understanding of data attributes and distribution, which are essential for the development of effective ML models.
In conclusion, the EDALearn benchmark suite is a significant step forward in the fusion of machine learning and EDA, providing a unified and systematic platform for performance evaluation, fostering collaboration and knowledge sharing, and encouraging innovation in ML-based algorithms and methodologies. We anticipate that this initiative will contribute substantially to the evolution and optimization of EDA tools and processes, steering the direction of future research in the field.

\begin{acks}
This work is supported by SRC 3104.001 and NSF 2106828.
\end{acks}

\bibliographystyle{ACM-Reference-Format}
\bibliography{references}

\end{document}